\documentclass[letterpaper]{article}
\def\year{2017}\relax

\usepackage{aaai17}
\usepackage{times}
\usepackage{helvet}
\usepackage{courier}
\usepackage{graphicx}
\usepackage{multirow}
\usepackage{url}
\usepackage{tablefootnote}

\begin{document}
%
\title{Twitter Job/Employment Corpus: \\ A Dataset of Job-Related Discourse Built with Humans in the Loop}
\author{Tong Liu, Christopher M. Homan\\
Golisano College of Computing and Information Sciences\\
Rochester Institute of Technology\\
{\tt tl8313@rit.edu, cmh@cs.rit.edu}\\
}
\maketitle
\begin{abstract}
We present the Twitter Job/Employment Corpus, a collection of tweets annotated by a humans-in-the-loop supervised learning framework that integrates crowdsourcing contributions and expertise on the local community and employment environment. Previous computational studies of job-related phenomena have used corpora collected from workplace social media that are hosted internally by the employers, and so lacks independence from latent job-related coercion and the broader context that an open domain, general-purpose medium such as Twitter provides. Our new corpus promises to be a benchmark for the extraction of job-related topics and advanced analysis and modeling, and can potentially benefit a wide range of research communities in the future.
\end{abstract}

\section{Introduction}

\noindent Working American adults spend more than one third of their daily time on job-related activities \cite{timeuse}---more than on anything else. Any attempt to understand a working individual's experiences, state of mind, or motivations must take into account their life at work. In the extreme, job dissatisfaction poses serious health risks and even leads to suicide \cite{occupationalsuicide,worksuicide}. 

Conversely, behavioral and mental problems greatly affect employee's productivity and loyalty. 70\% of US workers are disengaged at work \cite{gallup}. Each year lost productivity costs between 450 and 550 billion dollars. Disengaged workers are 87\% more likely to leave their jobs than their more satisfied counterparts are \cite{gallup}. The deaths by suicide among working age people (25-64 years old) costs more than \$44 billion annually \cite{suicidefigure}. 
By contrast, behaviors such as helpfulness, kindness and optimism predict greater job satisfaction and positive or pleasurable engagement at work \cite{harzer2013application}.

A number of computational social scientists have studied organizational behavior, professional attitudes, working mood and affect \cite{yardi2008pulse,kolari2007structure,brzozowski2009watercooler,de2013understanding}, but in each case: the data they investigated were collected from internal interactive platforms hosted by the workers' employers. 

These studies are valuable in their own right, but one evident limitation is that each dataset is limited to depicting a particular company and excludes the populations who have no access to such restricted networks (e.g., people who are not employees of that company). Moreover, the workers may be unwilling to express, e.g., negative feelings about work (``\emph{I don't wanna go to work today}''), unprofessional behavior (``\emph{Got drunk as hell last night and still made it to work}''), or a desire to work elsewhere (``\emph{I want to go work at Disney World so bad}'') on platforms controlled by their employers. 

A major barrier to studying job-related discourse on \textit{general-purpose}, \textit{public} social media---one that the previous studies did not face---is the problem of determining which posts are job-related in the first place. There is no authoritative training data available to model this problem. Since the datasets used in previous work were collected in the workplace during worktime, the content is implicitly job-related. By contrast, the subject matter of public social media is much more diverse. People with various life experiences may have different criteria for what constitutes a ``job'' and describe their jobs differently. 

For instance, a tweet like ``\emph{@SOMEONE @SOMEONE shit manager shit players shit everything}'' contains the job-related signal word ``manager,'' yet the presence of ``players'' ultimately suggests this tweet is talking about a sport team. Another example ``\emph{@SOMEONE anytime for you boss lol}'' might seem job-related, but ``boss'' here could also simply refer to ``friend'' in an informal and acquainted register. 



Extracting job-related information from Twitter can be valuable to a range of stakeholders. For example, public health specialists, psychologists and psychiatrists could use such first-hand reportage of work experiences to monitor job-related stress at a community level and provide professional support if necessary. Employers might analyze these data and use it to improve how they manage their businesses. 
It could help employees to maintain better online reputations for potential job recruiters as well. It is also meaningful to compare job-related tweets against non-job-related discourse to observe and understand the linguistic and behavioral similarities and differences between on- and off-hours.

Our main contributions are:

\begin{enumerate}
    \item We construct and provide a corpus of annotated tweets, the Twitter Job/Employment Corpus, which contains approximately 0.2 million job-related tweets and 6.8 million not-job-related tweets. To the best of our knowledge, we are the first to extract and study job-related discourse in general-purpose, public social media.
    \item We develop and improve an effective humans-in-the-loop classification framework for open-domain concepts such as job/employment that alternates between human annotation and automatic predictions by machine learning techniques over multiple iterations. This integrated mechanism largely reduces the human efforts in corpus annotation.
    \item We propose a qualified heuristic to separate business accounts from personal accounts relying on their linguistic styles and posts history. 
\end{enumerate}

\section{Background and Related Work}

Social media accounts for about 20\% of the time spent online \cite{socialtop}. Online communication can embolden people to reveal their cognitive state in a natural, un-self-conscious manner \cite{iKeepSafe}. Mobile phone platforms help social media to capture personal behaviors whenever and wherever possible \cite{de2013predicting,sadilek2013modeling}. These signals are often temporal, and can reveal how phenomena change over time. Thus, aspects about individuals or groups, such as preferences and perspectives, affective states and experiences, communicative patterns, and socialization behaviors can, to some degree, be analyzed and computationally modeled continuously and unobtrusively \cite{de2013predicting}.

Twitter has drawn much attention from researchers in various disciplines in large part because of the volume and granularity of publicly available social data associated with massive information. This micro-blogging website, which was launched in 2006, has attracted more than 500 million registered users by 2012, with 340 million tweets posted every day. Twitter supports directional connections (followers and followees) in its social network, and allows for geographic information about where a tweet was posted if a user enables location services. The large volume and desirable features provided by Twitter makes it a well-suited source of data for our task.

We focus on a broad discourse and narrative theme that touches most adults worldwide. Measures of volume, content, affect of job-related discourse on social media may help understand the behavioral patterns of working people, predict labor market changes, monitor and control satisfaction/dissatisfaction with respect to their workplaces or colleagues, and help people strive for positive change \cite{de2013understanding}. The language differences exposed in social media have been observed and analyzed in relation to location \cite{cheng2010you}, gender, age, regional origin, and political orientation \cite{rao2010classifying}. However, it is probably due to the natural challenges of Twitter messages --- conversational style of interactions, lack of traditional spelling rules, and 140-character limit of each message---we barely see similar public Twitter datasets investigating open-domain problems like job/employment in computational linguistic or social science field. Li et al. \shortcite{li2014major} proposed a pipelined system to extract a wide variety of major life events, including job, from Twitter. Their key strategy was to build a relatively clean training dataset from large volume of Twitter data with minimum human efforts. Their real world testing demonstrates the capability of their system to identify major life events accurately. The most parallel work that we can leverage here is the method and corpus developed by Liu et al. \shortcite{liu2016understanding}, which is an effective supervised learning system to detect job-related tweets from individual and business accounts. To fully utilize the existing resources, we build upon the corpus by Liu et al. \shortcite{liu2016understanding} to construct and contribute our more fine-grained corpus of job-related discourse with improvements of the classification methods. 

\section{Data and Methods}

\begin{figure}[ht]
\centering
\includegraphics[scale=0.6]{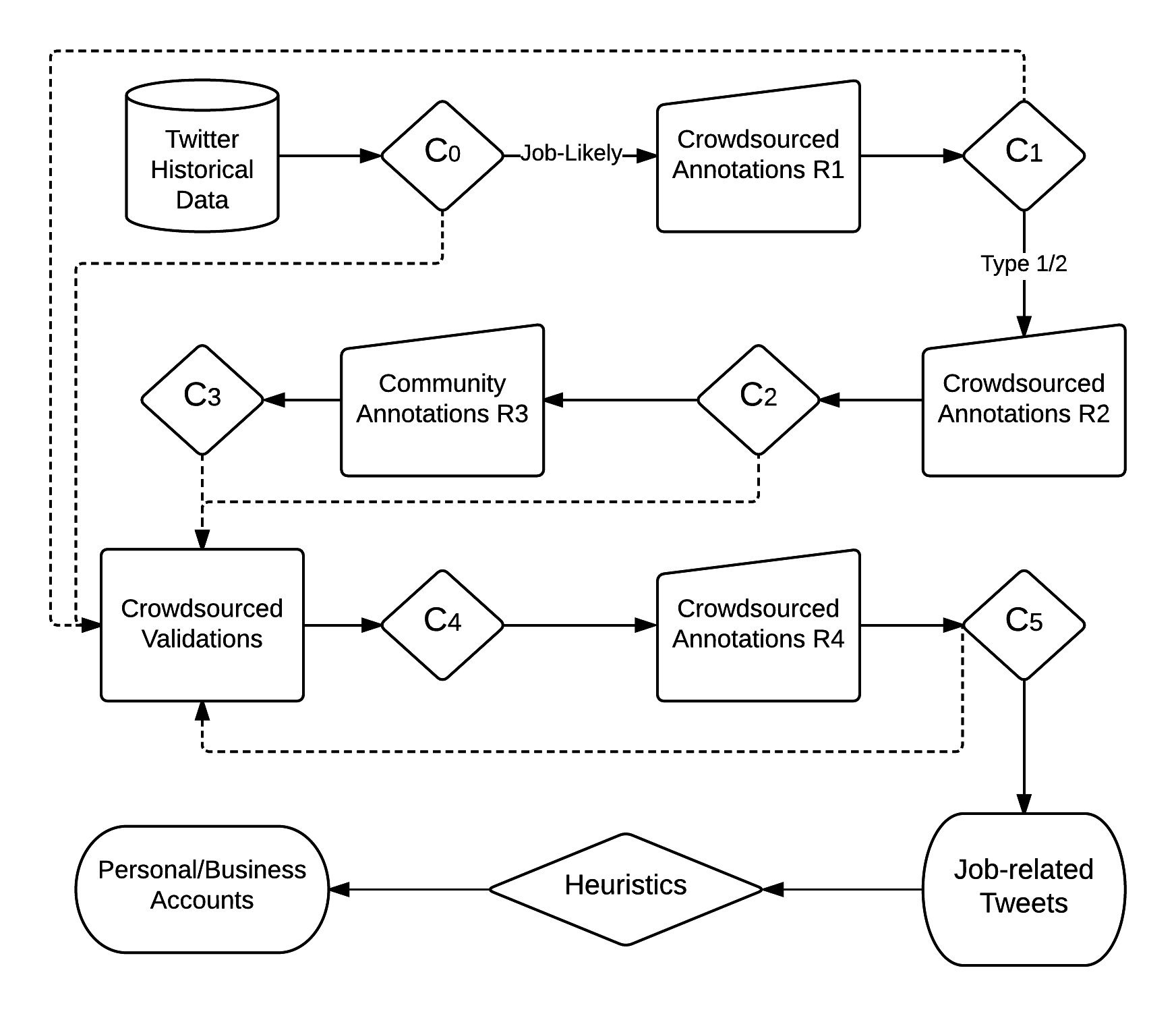}
\caption{Our humans-in-the-loop framework collects labeled data by alternating between human annotation and automatic prediction models over multiple rounds. Each diamond represents an automatic classifier (\textbf{C}), and each trapezoid represents human annotations (\textbf{R}). Each classifier filters and provides machine-predicted labels to tweets that are published to human annotators in the consecutive round. The human-labeled tweets are then used as training data by the succeeding automatic classifier. We use two types of classifiers: rule-based classifiers ($C_0$ and $C_4$) and support vector machines ($C_1$, $C_2$, $C_3$ and $C_5$). This framework serves to reduce the amount of human efforts needed to acquire large amounts of high-quality labeled data.}
\label{workflow}
\end{figure}

Figure \ref{workflow} shows the workflow of our humans-in-the-loop framework. It has multiple iterations of human annotations and automatic machine learning predictions, followed by some linguistic heuristics, to extract job-related tweets from personal and business accounts.

Compared to the framework introduced in \cite{liu2016understanding}, our improvements include: introducing a new rule-based classifier ($C_4$), conducting an additional round of crowdsourcing annotations (R4) to enrich the human labeled data, and training a classification model with enhanced performances ($C_5$) which was ultimately used to label the unseen data.

\subsection{Data Collection}

Using the DataSift\footnote{\url{http://datasift.com/}} Firehose, we collected historical tweets from public accounts with geographical coordinates located in a 15-counties region surrounding a medium sized US city 
from July 2013 to June 2014. This one-year data set contains over 7 million geo-tagged tweets (approximately 90\% written in English) from around 85,000 unique Twitter accounts.
This particular locality has geographical diversity, covering both urban and rural areas and providing mixed and balanced demographics. We could apply local knowledge into the construction of our final job-related corpus, which has been approved very helpful in the later experiments.

\subsection{Initial Classifier $\mathbf{C_0}$}

In order to identify probable job-related tweets which are talking about paid positions of regular employment while excluding noises (such as students discussing homework or school-related activities, or people complimenting others), we defined a simple term-matching classifier with inclusion and exclusion terms in the first step (see Table \ref{filter}).

Classifier $C_0$ consists of two rules: the matched tweet must contain at least one word in the \emph{Include} lexicon and it cannot contain any word in the \emph{Exclude} lexicon. Before applying filtering rules, we pre-processed each tweet by (1) converting all words to lower cases; (2) stripping out punctuation and special characters; and (3) normalizing the tweets by mapping out-of-vocabulary phrases (such as abbreviations and acronyms) to standard phrases using a dictionary of more than 5,400 slang terms in the Internet\footnote{\url{http://www.noslang.com/}}.

This filtering yielded over 40,000 matched tweets having at least five words, referred as \emph{job-likely}. 

\begin{table}[htbp]
\centering 
\begin{tabular}{c|c}
\multirow{2}{*}{\textbf{Include}} & job, jobless, manager, boss \\ 
 & my/your/his/her/their/at work \\ \hline
\multirow{2}{*}{\textbf{Exclude}} & school, class, homework, student, course \\ 
 & finals, good/nice/great job, boss ass\tablefootnote{Describe something awesome in a sense of utter dominance, magical superiority, or being ridiculously good.}  \\
\end{tabular}
\caption{The lexicons used by $C_0$ to extract the \emph{job-likely} set.}
\label{filter}
\end{table}

\subsection{Crowdsourced Annotation R1}

Our conjecture about crowdsourced annotations, based on the experiments and conclusions from \cite{snow2008cheap}, is that non-expert contributors could produce comparable quality of annotations when evaluating against those gold standard annotations from experts. And it is similarly effective to use the labeled tweets with high inter-annotator agreement among multiple non-expert annotators from crowdsourcing platforms to build robust models as doing so on expert-labeled data.

We randomly chose around 2,000 \emph{job-likely} tweets and split them equally into 50 subsets of 40 tweets each. In each subset, we additionally randomly duplicated five tweets in order to measure the intra-annotator agreement and consistency. We then constructed Amazon Mechanical Turk (AMT)\footnote{\url{https://www.mturk.com/mturk/welcome}} Human Intelligence Tasks (HITs) to collect reference annotations from crowdsourcing workers. We assigned 5 crowdworkers to each HIT---this is an empirical scale for crowdsourced linguistic annotation tasks suggested by previous studies \cite{callison2009fast,evanini2010using}. Crowdsourcing workers were required to live in the United States and had records of approval rating of 90\% or better. They were instructed to read each tweet and answer following question ``\emph{Is this tweet about job or employment?}'': their answer \texttt{Y} represents \emph{job-related} and \texttt{N} represents \emph{not job-related}. Workers were allowed to work on as many distinct HITs as they liked.

We paid each worker \$1.00 per HIT and gave extra bonuses to those who completed multiple HITs. We rejected workers who did not provide consistent answers to the duplicate tweets in each HIT. Before publishing the HITs to crowdsourcing workers, we consulted with Turker Nation\footnote{\url{http://www.turkernation.com}} to ensure that we treat and compensate workers fairly for their requested tasks.

Given the sensitive nature of this work, we anonymized all tweets to minimize any inadvertent disclosure of personal information ($@$names) or cues about an individual’s online identity (URLs) before publishing tweets to crowdsourcing workers. We replaced $@$names with \emph{$@SOMEONE$}, and recognizable URLs with \emph{$HTTP://LINK$}. No attempt was ever made to contact or interact with any user.

This labeling round yielded 1,297 tweets labeled with unanimous agreement among five workers, i.e. five workers gave the same label to one tweet---1,027 of these were labeled \emph{job-related}, and the rest 270 were \emph{not job-related}. They composed the first part of our human-annotated dataset, named as \textbf{Part-1}. 

\subsection{Training Helper Labeler $\mathbf{C_1}$}



\subsubsection{Feature Preparation}

We relied on the textual representations---a feature space of n-grams (unigrams, bigrams and trigrams)---for training. Due to the noisy nature of Twitter, where users frequently write short, informal spellings and grammars, we pre-processed input data as the following steps: (1) utilized a revised \texttt{Twokenizer} system which was specially trained on Twitter texts \cite{owoputi2013improved} to tokenize raw messages, (2) completed stemming and lemmatization using WordNet Lemmatizer \cite{bird2009natural}.

\subsubsection{Parameter Selection}

Considering the class imbalance situations in the training dataset, we selected the optimal learning parameters by grid-searching on a range of class weights for the positive (job-related) and negative (not job-related) classes, and then chose the estimator that optimized F1 score, using 10-fold cross validation. 

\subsubsection{First Helper $C_1$}

In \emph{Part-1} set, there are 1,027 job-related and 270 not job-related tweets. To construct a balanced training set for $C_1$, we randomly chose 757 tweets outside the \emph{job-likely} set (which were classified as negative by $C_0$). Admittedly these additional samples do not necessarily represent the true negative tweets (\emph{not job-related}) as they have not been manually checked. The noise introduced into the framework would be handled by the next round of crowdsourced annotations.

We trained our first SVM classification model \textbf{$C_1$} and then used it to label the remaining data in our data pool. 

\subsection{Crowdsourced Annotation R2}

We conducted the second round of labeling on a subset of $C_1$-predicted data to evaluate the effectiveness of the aforementioned helper $C_1$ and collect more human labeled data to build a class-balanced set (for training more robust models).

After separating positive- and negative-labeled (\emph{job-related} vs. \emph{not job-related}) tweets, we sorted each class in descending order of their confidence scores. We then spot-checked the tweets to estimate the frequency of job-related tweets as the confidence score changes. We discovered that among the top-ranked tweets in the positive class about half, and near the separating hyperplane (i.e., where the confidence scores are near zero) almost none, are truly job-related.

We randomly selected 2,400 tweets from those in the top 80th percentile of confidence scores in positive class (\emph{Type-1}). The \emph{Type-1} tweets are automatically classified as positive, but some of them may not be job-related in the ground truth. Such tweets are the ones which $C_1$ fails though $C_1$ is very confident about it. We also randomly selected about 800 tweets from those tweets having confidence scores closest to zero approaching from the positive side, and another 800 tweets from the negative side (\emph{Type-2}). These 1,600 tweets have very low confidence scores, representing those $C_1$ cannot clearly distinguish. Thus the automatic prediction results of the \emph{Type-2} tweets have a high chance being wrongly predicted. 
Hence, we considered both the clearer core and at the gray zone periphery of this meaningful phenomenon. 

Crowdworkers again were asked to annotate this combination of \emph{Type-1} and \emph{Type-2} tweets in the same fashion as in R1. Table \ref{round2_model_human} records annotation details.

\begin{table}
\centering
\begin{tabular}{c|c|c|c||c|c|c}
\multirow{3}{*}{\textbf{R2}} & \multicolumn{6}{c}{\textbf{\begin{tabular}[c]{@{}c@{}}Number of agreements\\  among 5 annotators\end{tabular}}} \\ \cline{2-7} 
 & \multicolumn{3}{c||}{\textbf{job-related}} & \multicolumn{3}{c}{\textbf{not job-related}} \\ \cline{2-7} 
 & \textbf{3} & \textbf{4} & \textbf{5} & \textbf{3} & \textbf{4} & \textbf{5} \\ \hline
\textbf{Type-1} & 129 & 280 & \textbf{713} & 50 & 149 & \textbf{1,079} \\ \hline
\textbf{Type-2} & 11 & 7 & \textbf{8} & 16 & 67 & \textbf{1,489} \\ 
\end{tabular}
\caption{Summary of annotations in R2 (showing when 3 / 4 / 5 of 5 annotators agreed).}
\label{round2_model_human}
\end{table}

Grouping \emph{Type-1} and \emph{Type-2} tweets with unanimous labels in R2 (bold columns in Table \ref{round2_model_human}), we had our second part of human-labeled dataset (\textbf{Part-2}). 

\subsection{Training Helper Labeler $\mathbf{C_2}$}

Combining \emph{Part-1} and \emph{Part-2} data into one training set---4,586 annotated tweets with perfect inter-annotator agreement (1748 job-related tweets and 2838 not job-related), we trained the machine labeler $C_2$ similarly as how we obtained $C_1$. 

\subsection{Community Annotation R3}

Having conducted two rounds of crowdsourced annotations, we noticed that crowdworkers could not reach consensuses on a number of tweets which were not unanimously labeled. This observation intuitively suggests that non-expert annotators inevitably have diverse types of understanding about the job topic because of its subjectivity and ambiguity. Table \ref{samples} provides examples (selected from both R1 and R2) of tweets in six possible inter-annotator agreement combinations. 

\begin{table}
\centering
\begin{tabular}{c|c}
\textbf{\begin{tabular}[c]{@{}c@{}}Crowdsourced\\ Annotations\\ Y/N\end{tabular}} & \textbf{Sample Tweet} \\ \hline
\textbf{Y Y Y Y Y} & \begin{tabular}[c]{@{}c@{}}Really bored....., no entertainment\\ at work today\end{tabular} \\ \hline
\textbf{Y Y Y Y N} & \begin{tabular}[c]{@{}c@{}}two more days of work then\\ I finally get a day off.\end{tabular} \\ \hline
\textbf{Y Y Y N N} & \begin{tabular}[c]{@{}c@{}}Leaving work at 430 and\\ driving in this snow is going\\ to be the death of me\end{tabular} \\ \hline
\textbf{Y Y N N N} & \begin{tabular}[c]{@{}c@{}}Being a mommy is the hardest\\ but most rewarding job\\ a women can have\\ \#babyBliss \#babybliss\end{tabular} \\ \hline
\textbf{Y N N N N} & \begin{tabular}[c]{@{}c@{}}These refs need to\\ DO THEIR FUCKING JOBS\end{tabular} \\ \hline
\textbf{N N N N N} & \begin{tabular}[c]{@{}c@{}}One of the best Friday nights\\ I've had in a while\end{tabular} \\ 
\end{tabular}
\caption{Inter-annotator agreement combinations and sample tweets.}
\label{samples}
\end{table}

Two experts from the local community with prior experience in employment were actively introduced into this phase to review tweets on which crowdworkers disagreed and provided their labels. The tweets with unanimous labels in two rounds of crowdsourced annotations were not re-annotated by experts because unanimous votes are hypothesized to be reliable as experts' labels. Table \ref{community} records the numbers of tweets these two community annotators corrected.


We have our third part of human-annotated data (\textbf{Part-3}): tweets reviewed and corrected by the community annotators.

\begin{table}
\centering
\begin{tabular}{c|c|c}
\textbf{R1 + R2} & \textbf{job-related} & \textbf{not job-related} \\ \hline
\textbf{Y Y Y Y N} & 644 & 5 \\ 
\textbf{Y Y Y N N} & 185 & 17 \\ 
\textbf{Y Y N N N} & 57 & 51 \\ 
\textbf{Y N N N N} & 11 & 301 \\ 
\hline
\textbf{Total} & 897 & 374
\end{tabular}
\caption{Summary of R3 community-based reviewed-and-corrected annotations.}
\label{community}
\end{table}

\subsection{Training Helper Labeler $\mathbf{C_3}$}

Combining \emph{Part-3} with all unanimously labeled data from the previous rounds (\emph{Part-1} and \emph{Part-2}) yielded 2,645 gold-standard-labeled job-related and 3,212 not job-related tweets. We trained $C_3$ on this entire training set. 

\subsection{Crowdsourced Validation of $\mathbf{C_0}$, $\mathbf{C_1}$, $\mathbf{C_2}$ and $\mathbf{C_3}$}

These three learned labelers ($C_1$, $C_2$, and $C_3$) are capable to annotate unseen tweets automatically. Their performances may vary due to the progressively increasing size of training data. 

To evaluate the models in different stages uniformly---including the initial rule-based classifier $C_0$---we adopted a post-hoc evaluation procedure: We sampled 400 distinct tweets that have not been used before from the data pool labeled by $C_0$, $C_1$, $C_2$ and $C_3$ respectively (there is no intersection between any two sets of samples). We had these four classifiers to label this combination of 1600-samples test set. We then asked crowdsourcing workers to validate a total of 1,600 unique samples just like our settings in previous rounds of crowdsourced annotations (R1 and R2). We took the majority votes (where at least 3 out of 5 crowdsourcing workers agreed) as reference labels for these testing tweets.


Table \ref{crowdsourced_validation} displays the classification measures of the predicted labels as returned by each model against the reference labels provided by crowdsourcing workers, and shows that $C_3$ outperforms $C_0$, $C_1$ and $C_{2}$.

\begin{table}[htbp]
\centering
\begin{tabular}{c|c|c|c|c}
\textbf{Model} & \textbf{Class} & \textbf{P} & \textbf{R} & \textbf{F1} \\ \hline
\multirow{3}{*}{$\mathbf{C_0}$} & job & 0.72 & 0.33 & 0.45 \\ 
 & notjob & 0.68 & 0.92 & 0.78 \\ 
 & \textit{avg / total} & \textit{0.70} & \textit{0.69} & \textit{0.65} \\ \hline
\multirow{3}{*}{$\mathbf{C_1}$} & job & 0.79 & 0.82 & 0.80 \\ 
 & notjob & 0.88 & 0.86 & 0.87 \\ 
 & \textit{avg / total} & \textit{0.85} & \textit{0.84} & \textit{0.84} \\ \hline
\multirow{3}{*}{$\mathbf{C_2}$} & job & 0.82 & 0.95 & 0.88 \\ 
 & notjob & 0.97 & 0.86 & 0.91 \\ 
 & \textit{avg / total} & \textit{0.91} & \textit{0.90} & \textit{0.90} \\ \hline
\multirow{3}{*}{$\mathbf{C_3}$} & job & 0.83 & 0.96 & 0.89 \\ 
 & notjob & 0.97 & 0.87 & 0.92 \\ 
 & \textit{avg / total} & \textit{0.92} & \textit{0.91} & \textit{0.91} \\ 
\end{tabular}
\caption{Crowdsourced validations of samples identified by models $C_0$, $C_1$, $C_2$ and $C_3$.}
\label{crowdsourced_validation}
\end{table}

\subsection{Crowdsourced Annotation R4}

Even though $C_3$ achieves the highest performance among four, it has scope for improvement. We manually checked the tweets in the test set that were incorrectly classified as \emph{not job-related} and focused on the language features we ignored in preparation for the model training. After performing some pre-processing on the tweets in false negative and true positive groups from the above testing phase, we ranked and compared their distributions of word frequencies. These two rankings reveal the differences between the two categories (false negative vs. true positive) and help us discover some signal words that were prominent in false negative group but not in true positive---if our trained models are able to recognize these features when forming the separating boundaries, the prediction false negative rates would decrease and the overall performances would further improve.

Our fourth classifier $C_4$ is rule-based again and to extract more potential job-related tweets, especially those would have been misclassified by our trained models. The lexicons in $C_4$ include the following signal words: \emph{career}, \emph{hustle}, \emph{wrk}, \emph{employed}, \emph{training}, \emph{payday}, \emph{company}, \emph{coworker} and \emph{agent}. 

We ran $C_4$ on our data pool and randomly selected about 2,000 tweets that were labeled as positive by $C_4$ and never used previously (i.e., not annotated, trained or tested in $C_0$, $C_1$, $C_2$, and $C_3$). We published these tweets to crowdsouring workers using the same settings of R1 and R2. The tweets with unanimously agreed labels in R4 form the last part of our human-labeled dataset (\textbf{Part-4}).

Table \ref{AMT_rounds} summarizes the results from multiple crowdsourced annotation rounds (R1, R2 and R4). 


\begin{table}[htbp]
\centering
\begin{tabular}{c|c|c|c||c|c|c}
\multirow{3}{*}{\textbf{Round}} & \multicolumn{6}{c}{\textbf{\begin{tabular}[c]{@{}c@{}}Number of agreements\\ among 5 annotators\end{tabular}}} \\ \cline{2-7} 
 & \multicolumn{3}{c||}{\textbf{job-related}} & \multicolumn{3}{c}{\textbf{not job-related}} \\ \cline{2-7} 
 & \textbf{3} & \textbf{4} & \textbf{5} & \textbf{3} & \textbf{4} & \textbf{5} \\ \hline
\textbf{R1} & 104 & 389 & 1,027 & 82 & 116 & 270 \\ \hline
\textbf{R2} & 140 & 287 & 721 & 68 & 216 & 2,568 \\ \hline
\textbf{R4} & 214 & 192 & 338 & 317 & 414 & 524 \\ 
\end{tabular}
\caption{Summary of crowdsourced annotations (R1, R2 and R4).}
\label{AMT_rounds}
\end{table}

\subsection{Training Labeler $\mathbf{C_5}$}

Aggregating separate parts of human-labeled data (\emph{Part-1} to \emph{Part-4}), we obtained an integrated training set with 2,983 job-related tweets and 3,736 not job-related tweets and trained $C_5$ upon it. We tested $C_5$ using the same data in crowdsourced validation phase (1,600 tested tweets) and discovered that $C_5$ beats the performances of other models (Table \ref{crowdsourced_validation_C5}). 

\begin{table}[htbp]
\centering
\begin{tabular}{c|c|c|c|c}
\textbf{Model} & \textbf{Class} & \textbf{P} & \textbf{R} & \textbf{F1} \\ \hline
\multirow{3}{*}{\textbf{$C_5$}} & job & 0.83 & 0.97 & 0.89 \\ 
 & notjob & 0.98 & 0.87 & 0.92 \\ 
 & \textit{avg / total} & \textit{0.92} & \textit{0.91} & \textit{0.91} \\
\end{tabular}
\caption{Performances of $C_5$.}
\label{crowdsourced_validation_C5}
\end{table}

Table \ref{topfeatures} lists the top 15 features for both classes in $C_5$ with their corresponding weights. Positive features (job-related) unearth expressions about personal job satisfaction (\emph{lovemyjob}) and announcements of working schedules (\emph{day off}, \emph{break}) beyond our rules defined in $C_0$ and $C_4$. Negative features (not job-related) identify phrases to comment on others' work (\emph{your work}, \emph{amazing job}, \emph{awesome job}, \emph{nut job}) though they contain ``work'' or ``job,'' and show that school- or game-themed messages (\emph{college career}, \emph{play}) are not classified into the job class which meets our original intention.

\begin{table}[ht]
\centering
\begin{tabular}{c c||c c}
\textbf{job-related} & \textbf{weights} & \textbf{not job-related} & \textbf{weights} \\ \hline
job & 1.77 & your work & -0.61 \\ 
manager & 1.71 & like it & -0.60 \\ 
work & 1.69 & amazing job & -0.59 \\ 
wrk & 1.44 & did & -0.55 \\ 
payday & 1.23 & nut & -0.45 \\ 
my bos & 1.06 & nut job & -0.45 \\ 
jobs & 0.83 & bos as & -0.43 \\ 
lovemyjob & 0.81 & play & -0.41 \\ 
at work & 0.81 & awesome job & -0.38 \\ 
working & 0.75 & college career & -0.37 \\ 
my career & 0.74 & high & -0.36 \\ 
day off & 0.73 & doing & -0.35 \\ 
boss & 0.73 & hustle & -0.35 \\ 
service & 0.71 & you guy & -0.33 \\ 
break & 0.70 & love your & -0.33 
\end{tabular}
\caption{Top 15 features for both classes of \textbf{$C_5$}.}
\label{topfeatures}
\end{table}

\subsection{End-to-End Evaluation}

The class distribution in the machine-labeled test data is roughly balanced, which is not the case in real-world scenarios, where not-job-related tweets are much more common than job-related ones. 

We proposed an end-to-end evaluation: to what degree can our trained automatic classifiers ($C_1$, $C_2$, $C_3$ and $C_5$) identify job-related tweets in the real world? We introduced the \emph{estimated effective recall} under the assumption that for each model, the error rates in our test samples (1,600 tweets) are proportional to the actual error rates found in the entire one-year data set which resembles the real world. We labeled the entire data set using each classifier and defined the estimated effective recall $\hat{R}$ for each classifier as
\begin{eqnarray*}
\hat{R} &=& 
\frac{Y \cdot N_t \cdot R}{Y \cdot N_t \cdot R + N \cdot Y_t \cdot (1-R)}
\end{eqnarray*}
where $Y$ is the total number of the classifier-labeled job-related tweets in the entire one-year data set, $N$ is the total of not job-related tweets in the entire one-year data set, $Y_t$ is the number of classifier-labeled job-related tweets in our 1,600-sample test set, $N_t = 1,600 - Y_t$, and $R$ is the recall of the job class in our test set, as reported in Tables \ref{crowdsourced_validation} and \ref{crowdsourced_validation_C5}. 

\begin{table}[htbp]
\centering
\begin{tabular}{c|c|c|c|c}
\textbf{Models} & $\mathbf{C_1}$ & $\mathbf{C_2}$ & $\mathbf{C_3}$ & $\mathbf{C_5}$ \\ \hline
\textbf{Y} & 115,696 & 195,442 & 190,471 & 233,187 \\ \hline
\textbf{N} & 6,990,633 & 6,910,887 & 6,915,858 & 6,873,142 \\ \hline
\textbf{$\mathbf{Y_t}$} & 512 & 691 & 707 & 729 \\ \hline
\textbf{$\mathbf{N_t}$} & 1,088 & 909 & 893 & 871 \\ \hline
\textbf{R} & 0.82 & 0.95 & 0.96 & 0.97 \\ \hline
\textbf{$\mathbf{\hat{R}}$} & 0.14 & 0.41 & 0.46 & 0.57 \\ 
\end{tabular}
\caption{Estimated effective recalls for different trained models ($C_1$, $C_2$, $C_3$ and $C_5$) to identify job-related tweets in real world setting.}
\label{effective_recall}
\end{table}

Table \ref{effective_recall} shows that $C_5$ can be used as a good classifier to automatically label the topic of unseen data as job-related or not.

\subsection{Determining Sources of Job-Related Tweets}

Through observation we noticed some patterns like:

\begin{quotation}
\noindent
``\emph{Panera Bread: Baker - Night (\#Rochester, NY) HTTP://URL \#Hospitality \#VeteranJob \#Job \#Jobs \#TweetMyJobs}''
\end{quotation}

\noindent in the class of job-related tweets. Nearly every job-related tweet that contained at least one of the following hashtags: \emph{\#veteranjob}, \emph{\#job}, \emph{\#jobs}, \emph{\#tweetmyjobs}, \emph{\#hiring}, \emph{\#retail}, \emph{\#realestate}, \emph{\#hr} also had a URL embedded. We counted the tweets containing only the listed hashtags, and the tweets having both the queried hashtags and embedded URL, and summarized the statistics in Table \ref{hashtag_table}. By spot checking we found such tweets always led to recruitment websites. This observation suggests that these tweets with similar ``hashtags + URL'' patterns originated from business agencies or companies instead of personal accounts, because individuals by common sense are unlikely to post recruitment advertising. 

\begin{table}[htbp]
\centering
\begin{tabular}{c|c|cc}
\textbf{} & \textbf{hashtag only} & \textbf{hashtag + URL} & \textbf{\%} \\ \hline
\textbf{\#veteranjob} & 18,066 & 18,066 & 100.00\\ 
\textbf{\#job} & 79,359 & 79,326 & 99.96\\ 
\textbf{\#jobs} & 59,882 & 59,864 & 99.97\\ 
\textbf{\#tweetmyjobs} & 39,007 & 39,007 & 100.00\\ 
\textbf{\#hiring} & 622 & 619 & 99.52\\ 
\textbf{\#retail} & 17,107 & 17,105 & 99.99\\ 
\textbf{\#realestate} & 113 & 112 & 99.12\\ 
\textbf{\#hr} & 406 & 405 & 99.75
\end{tabular}
\caption{Counts of tweets containing the queried hashtags only, and their subsets of tweets with URL embedded.}
\label{hashtag_table}
\end{table}


This motivated a simple heuristic that appeared surprisingly effective at determining which kind of accounts each job-related tweet was posted from: if an account had more job-related tweets matching the ``hashtags + URL'' patterns than tweets in other topics, we labeled it a \emph{business} account; otherwise it is a \emph{personal} account. We validated its effectiveness using the job-related tweets sampled by the models in crowdsourced evaluations phase. It is essential to note that when crowdsourcing annotators made judgment about the type of accounts as \emph{personal} or \emph{business}, they were shown only one target tweet---without any contexts or posts history which our heuristics rely on.

Table \ref{source_validation} records the performance metrics and confirms that our heuristics to determine the sources of job-related tweets (\emph{personal} vs. \emph{business} accounts) are consistently accurate and effective.

\begin{table}[htbp]
\centering
\begin{tabular}{c|c|c|c|c}
\textbf{From} & \textbf{Class} & \textbf{P} & \textbf{R} & \textbf{F1} \\ \hline
\multirow{3}{*}{$\mathbf{C_1}$} & personal & 1.00 & 0.98 & 0.99 \\ 
 & business & 0.98 & 1.00 & 0.99 \\ 
 & \textit{avg/total} & \textit{0.99} & \textit{0.99} & \textit{0.99} \\ 
 \hline
\multirow{3}{*}{$\mathbf{C_2}$} & personal & 1.00 & 0.99 & 0.99 \\ 
 & business & 0.99 & 1.00 & 0.99 \\ 
 & \textit{avg/total} & \textit{0.99} & \textit{0.99} & \textit{0.99} \\ 
 \hline
\multirow{3}{*}{$\mathbf{C_3}$} & personal & 1.00 & 0.99 & 0.99 \\ 
 & business & 0.99 & 1.00 & 0.99 \\ 
 & \textit{avg/total} & \textit{0.99} & \textit{0.99} & \textit{0.99} \\ 
 \hline
\multirow{3}{*}{$\mathbf{C_5}$} & personal & 1.00 & 0.99 & 0.99 \\ 
 & business & 0.99 & 1.00 & 0.99 \\ 
 & \textit{avg/total} & \textit{0.99} & \textit{0.99} & \textit{0.99} \\ 
\end{tabular}
\caption{Evaluations of heuristics to determine the type of accounts (personal vs. business), job-related tweets sampled by different models in Table \ref{crowdsourced_validation}.}
\label{source_validation}
\end{table}

We used $C_5$ to detect (not) job-related tweets, and applied our linguistic heuristics to further separate accounts into personal and business groups automatically.

\section{Annotation Quality}

To assess the labeling quality of multiple annotators in crowdsourced annotation rounds (R1, R2 and R4), we calculated Fleiss' kappa \cite{fleiss1971measuring} and Krippendorff's alpha \cite{krippendorff2004content} measures using the online tool \cite{agreement_tool} to assess inter-annotator reliability among the five annotators of each HIT. And then we calculated the average and standard deviation of inter-annotator scores for multiple HITs per round. Table \ref{inter_annotator} records the inter-annotator agreement scores in three rounds of crowdsourced annotations.

\begin{table}[htbp]
\centering
\begin{tabular}{c|c|c}
\textbf{Round} & \textbf{Fleiss' kappa} & \textbf{Krippendorf's alpha} \\ \hline
\textbf{R1} & 0.62 $\pm$ 0.14 & 0.62 $\pm$ 0.14 \\ 
\textbf{R2} & 0.81 $\pm$ 0.09 & 0.81 $\pm$ 0.08 \\ 
\textbf{R4} & 0.42 $\pm$ 0.27 & 0.42 $\pm$ 0.27 \\ 
\end{tabular}
\caption{Inter-annotator agreement performance for our three rounds of crowdsourced annotations. Average $\pm$ stdev agreements are \emph{Good}, \emph{Very Good} and \emph{Moderate} \cite{altman1991inter} respectively.}
\label{inter_annotator}
\end{table}

The inter-annotator agreement between the two expert annotators from local community was assessed using Cohen's kappa \cite{cohen1960coefficient} as $\kappa = 0.803$ which indicates empirically almost excellent. Their joint efforts corrected more than 90\% of tweets which collected divergent labels from crowdsourcing workers in R1 and R2.

We observe in Table \ref{inter_annotator} that annotators in R2 achieved the highest average inter-annotator agreements and the lowest standard deviations than the other two rounds, suggesting that tweets in R2 have the highest level of confidence being related to job/employment. As shown in Figure \ref{workflow}, the annotated tweets in R1 are the outputs from $C_0$, the tweets in R2 are from $C_1$, and the tweets in R4 are from $C_4$. $C_1$ is a supervised SVM classifier, while both $C_0$ and $C_4$ are rule-based classifiers. The higher agreement scores in R2 indicate that a trained SVM classifier can provide more reliable and less noisy predictions (i.e., labeled data). Further, higher agreement scores in R1 than R4 indicates that the rules in $C_4$ are not intuitive as that in $C_1$ and introduce ambiguities.
For example, tweets ``\emph{What a career from Vince young!}'' and ``\emph{I hope Derrick Rose plays the best game of his career tonight}'' both use \emph{career} but convey different information: the first tweet was talking about this professional athlete's accomplishments while the second tweet was actually commenting on the game the user was watching. Hence crowdsourcing workers working on $C_4$ tasks read more ambiguous tweets and solved more difficult problems than those in $C_1$ tasks did. Considering that, it is not surprising that the inter-annotator agreement scores of R4 are the worst.

\section{Dataset Description}

Our dataset is available as a plain text file in JSON format. Each line represents one unique tweet with five attributes identifying the tweet id (\emph{tweet\_id}, a unique identification number generated by Twitter for each tweet), topics \emph{job} vs. \emph{notjob} labeled by human (\emph{topic\_human}) and machine (\emph{topic\_machine}), and sources \emph{personal} vs. \emph{business} labeled by human (\emph{source\_human}) and machine (\emph{source\_machine}). \emph{NA} represents ``not applicable.'' An example of tweet in our corpus is shown as follows:

\begin{verbatim}
{
    "topic_human":"NA",
    "tweet_id":"409834886405832705",
    "topic_machine":"job",
    "source_machine":"personal",
    "source_human":"NA"
}
\end{verbatim}


Table \ref{stats} provides the main statistics of our dataset w.r.t the topic and source labels provided by human and machine.


\begin{table}[htbp]
\centering
\begin{tabular}{c|c|c|c}
\multicolumn{2}{c|}{\textbf{Count of Labels}} & \textbf{Human} & \textbf{Machine} \\ \hline
\textbf{} & job & 2,978 & 233,187 \\ 
\textbf{Topic} & notjob & 3,736 & 6,873,142 \\ 
\textbf{} & NA & 842 & -- \\ \hline
\textbf{} & personal & 1,357 & 7,025,203 \\ 
\textbf{Source} & business & 232 & 81,126 \\ 
\textbf{} & NA & 5,966 & -- \\ 
\end{tabular}
\caption{Statistics of our dataset labeled by human and machine.}
\label{stats}
\end{table}

\subsubsection{Terms and Conditions}

According to the Twitter agreement and policy, we are allowed to only distribute tweet ids when providing downloadable datasets to third parties. This guarantees prompt response to the content changes reported through the Twitter API, such as deletions or the status changes (public/protected) of tweets\footnote{See more terms and conditions at \url{https://twitter.com/tos?lang=en}, and \url{https://dev.twitter.com/overview/terms/agreement-and-policy}.}.









\section{Conclusion}

We presented the Twitter Job/Employment Corpus and our  approach for extracting discourse on work from public social media.
We developed and improved an effective, humans-in-the-loop active learning framework that uses human annotation and automatic predictions over multiple rounds to label automatically data as job-related or not job-related. We accurately determine whether or not Twitter accounts are personal or business-related, according to their linguistic characteristics and posts history. Our crowdsourced evaluations suggest that these labels are precise and reliable. Our classification framework could be extended to other open-domain problems that similarly lack high-quality labeled ground truth data.

\bigskip

\bibliographystyle{aaai}
\bibliography{icwsm17.bib}

\end{document}